\documentclass{article}

     \usepackage[final]{neurips_2022}

\usepackage[utf8]{inputenc} 
\usepackage[T1]{fontenc}    
\usepackage{hyperref}       
\usepackage{url}            
\usepackage{booktabs}       
\usepackage{amsfonts}       
\usepackage{nicefrac}       
\usepackage{microtype}      

\usepackage{graphicx}
\usepackage[dvipsnames]{xcolor}
\usepackage{amsmath}
\usepackage{makecell}

\usepackage[export]{adjustbox}

\title{Power Grid Congestion Management \\ via Topology Optimization with AlphaZero}

\author{%
  Matthias Dorfer\thanks{Authors contributed equally.}   ,$\,$
  Anton R. Fuxjäger$^*\text{,}$
  Kristian Kozak$^*$,$\,$
  Patrick M. Blies,$\,$
  Marcel Wasserer \\
  enliteAI \\
  Vienna, A-1010 \\
  \texttt{m.wasserer@enlite.ai} \\
}

\begin{document}

%
%

\maketitle


\begin{abstract}

The energy sector is facing rapid changes in the transition towards clean renewable sources.
However, the growing share of volatile, fluctuating renewable generation such as wind or solar
energy has already led to an increase in power grid congestion and network security concerns.
Grid operators mitigate these by modifying either generation or demand
(redispatching, curtailment, flexible loads).
Unfortunately, redispatching of fossil generators leads to excessive grid operation costs and higher emissions,
which is in direct opposition to the decarbonization of the energy sector.
In this paper, we propose an AlphaZero-based grid topology optimization agent
as a non-costly, carbon-free congestion management alternative.
Our experimental evaluation confirms the potential of topology optimization for power grid operation, achieves a reduction of the average amount of required redispatching by 60\%, and shows the interoperability with traditional congestion management methods.
Our approach%
\footnote{Our submission is available at: \url{https://github.com/enlite-ai/maze-l2rpn-2022-submission}}
also ranked 1st in the WCCI 2022 Learning to Run a Power Network (L2RPN) competition%
\footnote{Learning to Run a~Power Network, \cite{marot2019l2rpn,marot2021retrospective,marot2021l2rpntrust}, \url{https://l2rpn.chalearn.org/}}.
Based on our findings, we identify and discuss open research problems
as well as technical challenges for a productive system on a real power grid.


\end{abstract}

\section{Introduction}
\label{sec:introduction}


Interconnected electrical power grids are a critical infrastructure of modern society. The operation of these grids is a demanding control task, requiring uninterrupted monitoring by skilled experts and frequent interventions to safely and reliably transport electricity from the producers (generators) to all connected consumers (loads)~\citep{Kelly2020gridrl}. Load flow calculation is a well-established method for the simulation of grid states and it is routinely used by grid operators to ensure that all security constraints are met~\citep{stott1974loadflow}. The existence of fast and reliable simulators pave the way for powerful tools based on Reinforcement Learning (RL)~\citep{marot2019l2rpn}, supporting operators with the increasingly difficult challenges in the ongoing effort to decarbonize the energy sector and ensure security of supply~\citep{marott2022towards,marot2022perspectives}.

\begin{figure}[t]
  \centerline{\includegraphics[width=.43\textwidth]{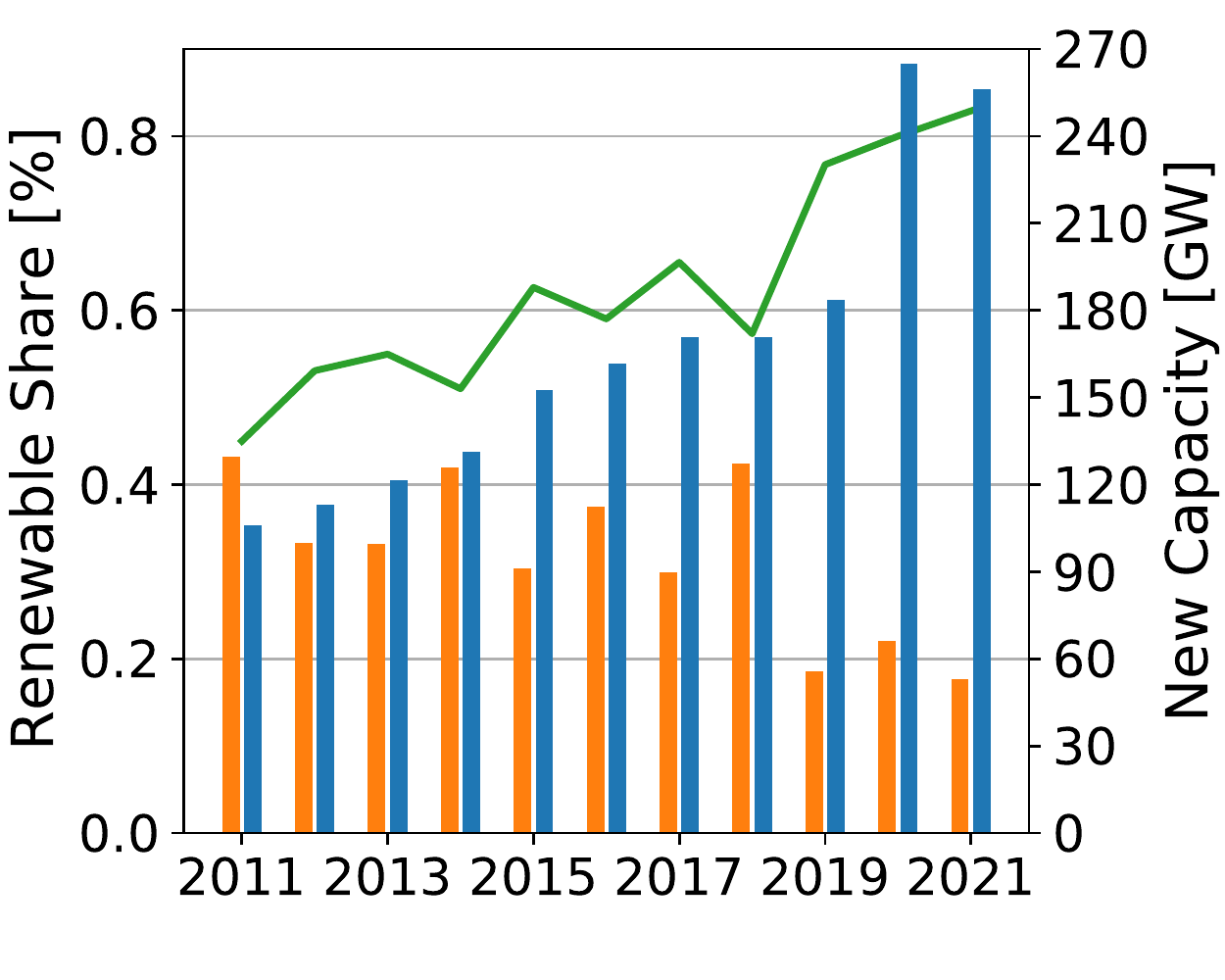}\includegraphics[width=.57\textwidth]{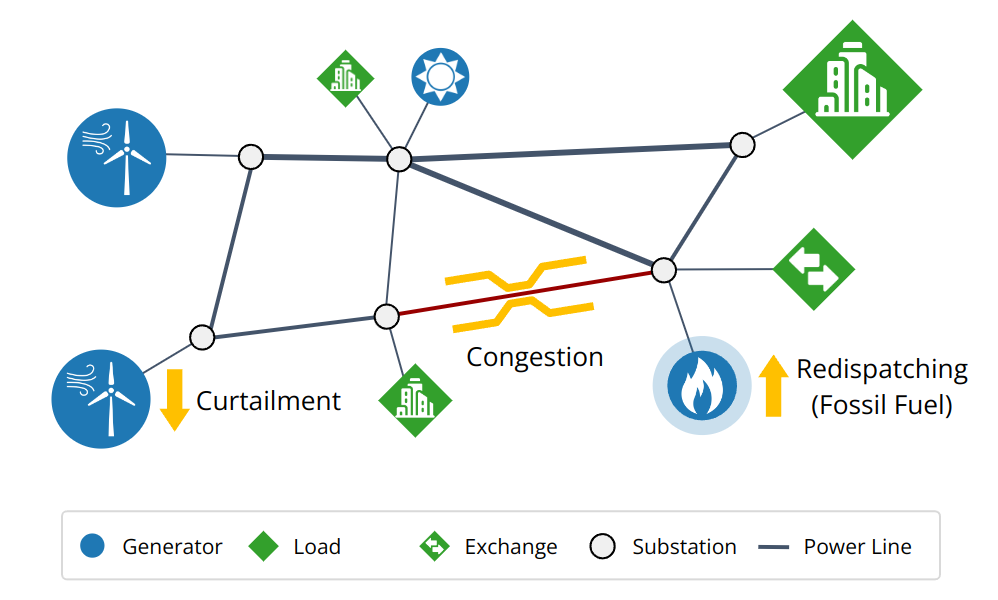}}
  \caption{Left: Total global generator capacity expansion per year (orange: non-renewables, blue: renewables, green: share of new renewable generation)
Right: Simplified congestion scenario with surplus power from wind farms and transmission to large remote consumers and a neighbouring country. Load flow analysis reveals a congested transmission line. Curtailing wind power and compensating for the spilled renewable energy by ramping up a thermal generator is the prevalent remedial action applied in practice.}
  \label{fig:renewables_and_congestion}
\end{figure}

The shift towards renewable electricity generation is advancing rapidly. Globally, 80\% of the newly installed generation capacity is already based on renewable sources, where solar and wind account for the largest share\footnote{Data provided by the International Renewable Energy Agency, \href{https://pxweb.irena.org/pxweb/en/IRENASTAT}{IRENASTAT}, retrieved Sept 2022.} (see Figure~\ref{fig:renewables_and_congestion}, left). Compared to non-renewable technologies such as thermal generation from fossil fuel, solar and wind energy is highly volatile, confronting grid operators with large power gradients and increased uncertainties. This makes more frequent interventions necessary in order to prevent individual grid elements from exceeding operational limits (grid congestion)~\citep{han2015congcasestudy}.
Figure~\ref{fig:renewables_and_congestion}, right shows an exemplary congestion scenario. 

\textbf{Power Grid Basics.}
Laying out the electrophysical foundations and covering all common grid elements
is beyond the scope of this paper.
Fortunately, this complexity is largely encapsulated in the simulation tools of the grid operators. A more detailed presentation is provided in~\citep{Kelly2020gridrl}.

\textit{Grid model}. In essence, a power grid connects generators (producers) to loads (consumers) and is modeled as an undirected multigraph,
with edges representing transmission lines (carrying energy over long distances) and nodes representing substations (switchable connection points)\footnote{Throughout the paper we adopt the view of a transmission grid operator, but note that the same concepts apply to the distribution grid as well.}.
Relevant grid elements are presented in more detail in Section~\ref{sec:env_disign}. 

\textit{Congestion management}. Given an expected load and generation scenario,
grid security analysis is conducted as part of the operational planning process~\citep{shahidehpour2005gridsecurity}. If the security analysis reveals violations of operational constraints,
grid operators are required to intervene to keep the grid stable and reliable.

\textit{Costly and non-costly measures}. Operators can relieve congested lines by altering the power entering and leaving the grid, which involves high costs from modifying previously agreed transmissions and procuring energy from different generators~\citep{VANDENBERGH2015redispatching}. From an economic perspective, it is highly desirable to jointly optimize these costly measures with the non-costly measure of dynamically reconfiguring the grid by making use of existing switchable equipment~\citep{hedman2011toporeview}.

\textbf{Related Work.}
Optimal line switching was first proposed by~\citet{fisher2008topo} as a mixed linear integer program, based on a linear approximation of the load flow calculation (DC formulation). \citet{hedman2009lineswitching} added security analysis and a heuristic strategy to obtain approximate results for the IEEE 118 bus system.
Note that line switching does not make use of the flow rerouting ability of the substations. Subsequent work~\citep{heidarifar2016topo, xiao2018topo, li2019topo, zhou2021topo} includes more complete modeling of substations at the circuit breaker level, but continues to rely on linear approximations of the grid dynamics in order to apply commercial solvers.
Recent editions of the Learning to run a Power Network Challenge~\citep{marot2021retrospective,marot2021l2rpntrust}
reveal the first promising results of RL-based topology optimizers.
\citet{zhou2021es} propose to combine search-based planning with the predictions of a policy network
learned via large scale Evolutionary Strategy (ES) training.
Their approach is able to discover long-term strategies,
but struggles with the inherent limitation of being sample inefficient.
\citet{roychowdhury2022toporl} train a value network with Dueling DQN (DDQN)
for line switching and topology control at a substation level.
The Authors then evaluate their approach on the IEEE 14 bus system. 
Using the same testbed~\citet{subramanian2021simplerl} approached the problem
by utilizing a cross-entropy optimizer for topological switching.
By directly predicting the desired target topology and subsequently planning an action sequence towards this target topology
~\citet{Yoon2021afterstate} propose an off-policy actor-critic approach
adopting the hierarchical policy together with the afterstate representation.
Related to our approach,~\citet{taha2022mcts} propose to learn a line load dynamics model, which is then
utilized in a tree search for planning viable topological remedial actions.
However, this tree search is not yet guided by a learned policy network,
making it inefficient and unlikely to find optimal solutions for upcoming congestion situations
within the~exponential topology exploration space.

\textbf{Our Contributions}.
We pick up this line of research and propose an AlphaZero-based topology optimization agent~\citep{silver2017alphazero}
to address power grid congestion management.
Our choice is motivated by recent successes of AlphaZero
for similar large-scale combinatorial optimization problems.
We benchmark our topology agent on a 118 substation power grid
and demonstrate how to successfully combine it with traditional congestion management measures.
Finally, we discuss the~practical relevance of such an agent
using the examples of day-ahead planning and real-time remedial action recommendation,
and outline promising further research directions.

\section{Environment Design and Interaction Workflow}
\label{sec:env_disign}
This section describes the power grid environment, its observation and action spaces, reward as well as custom modules.

\subsection{Power Grid Environment Overview}
\label{subsec:power_grid}

We utilize the Grid2Op framework~\citet{grid2op} provided by RTE France, Europe's largest grid operator. A power grid consists of generators, loads, storages, substations, and power lines. We provide a brief overview below; please refer to the \href{https://grid2op.readthedocs.io/en/latest/}{official documentation} for the full specification.

\begin{figure}[h]
  \centerline{\includegraphics[width=0.99\textwidth]{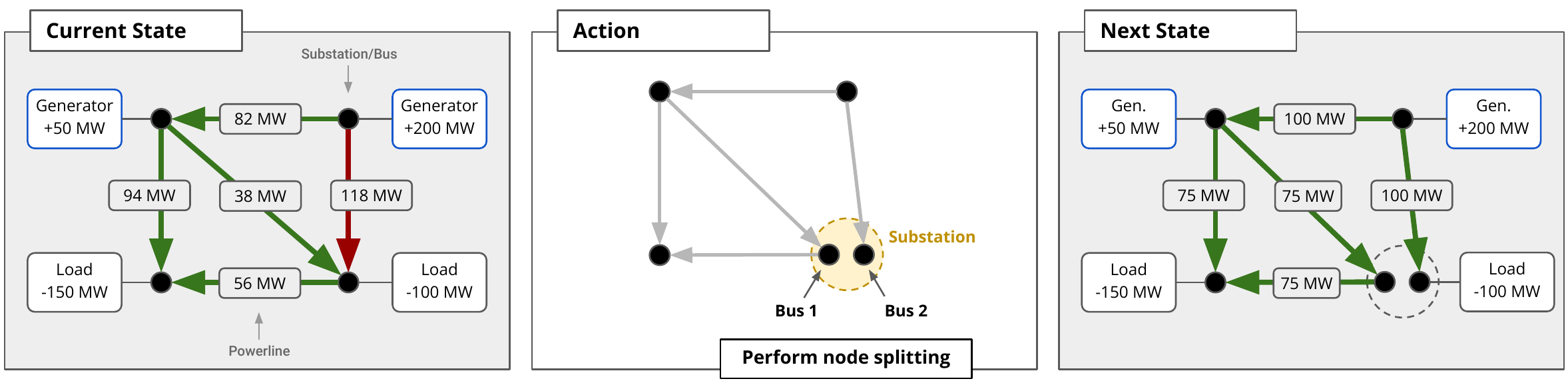}}
  \caption{Example topology action. Node splitting is performed by switching the overflowing line to the second bus, changing the topology of the network and resolving the overflow. Note that in the Grid2Op framework~\citet{grid2op}, each substation features two buses. In the real world, a~substation might feature even more than two buses.
  (Figure adopted from~\citep{marot2021retrospective}.)}
  \label{fig:topology_action_example}
\end{figure}

\textbf{Generators, Loads and Storages.} \emph{Generators} produce power. Renewable generators (sun, wind, hydro) follow simulated weather patterns; thermic generators (coal, gas, nuclear) follow predefined output curves. As a congestion management measure, the output of generators can be influenced---thermics can be redispatched (up or down) and renewables curtailed (down only). \emph{Loads} consume power, following a predefined pattern unknown to the agent. \emph{Storages} store power across time steps, and can be gradually charged or discharged.

\textbf{Transmission: Substations and Lines.} \emph{Substations} connect elements (generators, loads, lines), and consist of two buses. All connected elements need to be assigned to a bus, configuring the topology of the grid. For an example of how bus switching affects the topology of a network, refer to Figure~\ref{fig:topology_action_example}. \emph{Power lines} connect substations. Each line has a thermal limit (i.e., capacity), which it can exceed only for a small number of time steps---then an automated disconnect occurs. Severe overflows (>~200~\%) cause an immediate disconnect, and often lead to cascading line failures.

\textbf{Objective.} The goal is to keep the grid in operation, avoiding congestions and line disconnects
that can lead to isolation of loads or generators---i.e., a diverging power flow (black out).
The secondary objective is to minimize operational costs incurred by power loss and congestion management.

\subsection{Observation Space}
\label{subsec:observation_space}

\textbf{Overview.} We utilize the \texttt{CompleteObservation}\footnote{Please refer to the documentation of \texttt{CompleteObservation} for a complete description (including, e.g., maintenance and cool down times): \url{https://grid2op.readthedocs.io/en/latest/observation.html}}, which we further process
into feature representations.
The observation space includes (most notably, but not limited to):
\textit{generator and loads state}, including power production/consumption and its forecasts;
\textit{power line state}, including the power flow at origin and extremity, and the~current load as a fraction of capacity (denoted as~$\rho$, one of the most important metrics);
\textit{topology configuration}, with the power lines status and bus configurations for each element; and, finally, 
\textit{redispatch, curtailment and storage states}.

\textbf{Simulation.} A limited simulation functionality for planning purposes is available to the~agent. Compared to the complete power grid environment (which is not available to the agent), it features only forecasts of future load, generator, and weather patterns. Hence, while the agent can test the~general impact of an~action, the accuracy varies.


\subsection{Action Space and Combinatorial Complexity}
\label{subsec:action_space}

An action can modify topology, redispatching, curtailment and storage elements\footnote{For a more information, please review the action documentation: \url{https://grid2op.readthedocs.io/en/latest/action.html} and operational cost overview: \url{https://codalab.lisn.upsaclay.fr/competitions/5410\#learn_the_details-evaluation}}.

\textbf{Non-Costly Measures.} \textit{Topology actions} (the only non-costly measure) modify the bus configuration of one~grid element (either all connections on one line or one substation). As each substation features two buses, there are overall $2^{k}$ possible topological configurations, where $k$ is the number of connections in the grid (i.e., all elements on all substations). While not all configurations are valid,
this still yields an exponential exploration space, presenting a challenge for both RL-based and traditional approaches.

\textbf{Costly Measures.} \emph{Redispatching actions} (incurring redispatching costs) modify the target output of the redispatchable generators (up or down), and need to be within the per-generator bounds. \emph{Curtailment actions} (incurring curtailment costs) limit the production of renewable generators. \emph{Storage actions} (incurring loading, preserving, and discharging costs) direct the behavior of storages.

Note that in a real grid, a few more congestion remediation measures are available, such as counter trading, demand-side management or load shedding. However, the actions described above (and simulated in the Grid2Op environment) represent the main remediation options at hand.

\textbf{Action Space.} Our topology optimization agent utilizes the unitary action space provided by Grid2Op. This space represents all individual available topology actions (i.e., manipulating all combinations of bus connections, per grid element) as integers, creating a discrete action space. To reduce the action space size, we then perform action masking and action reduction (see Section~\ref{sec:experiments}).

\subsection{Environment Dynamics, Interaction Workflow and Reward}
\label{subsec:env_dynamics}

This section describes custom modules engaging in observation processing and environment control.

\textbf{Grid State Observer and Safe-State Skipping.} In most time steps, the grid is running safely, and launching our congestion remediation agent is neither necessary nor desirable (due to additional operational costs). Hence, we integrate a \emph{Grid State Observer} that determines whether the current state is considered safe by evaluating the current maximum line load\footnote{The \emph{Grid State Observer} is a customizable module and it can consider also, e.g., an n-1 contingency analysis.}.
In safe states, the congestion remediation agent is skipped, and either a recovery (see below) or a no-operation action is performed.

\textbf{Topology Recovery.} In the default topology, all connections are switched to the same bus, i.e., no node splitting takes place (see Figure~\ref{fig:topology_action_example} for a node splitting explanation). We find that this ,,fully connected`` setting is rather resilient, and a good starting point for congestion management. However, due to the reduced action set (see Section~\ref{sec:experiments}), the agent does not always have all the actions available required to recover this topology. Hence, in safe states, we perform automated topology recovery.


\textbf{Rewards.} In case of a congestion, the topology optimization agent is triggered. We utilize a shaped reward based on the cumulative sum of all overflowing line loads, which the agent aims to minimize.

\section{Topology Optimization with AlphaZero and Heuristic Value Functions}
\label{sec:alpha_zero}
%
%
In this section, we outline our customized AlphaZero algorithm employed with a heuristic value function. AlphaZero~\citep{silver2017alphazero}
constitutes an RL algorithm using a general-purpose Monte Carlo Tree Search (MCTS). Each search consists of a fixed number of simulations, starting at the root node, and traversing the tree until a new leaf node is added to the tree. Which action is selected at each branching in the tree is guided by the probabilistic upper confidence tree (PUCT) bound~\citep{rosin2011multi} taking into account the visitation count of the current and potential child node, the action probability sampled from the policy network and the value assigned to the resulting state when performing the respective action. Once the desired number of simulations has been reached, the action to perform is sampled from all expanded children of the root node w.r.t. the probability distribution obtained by taking the softmax of their visitation counts. In the greedy setting, the action with the highest visitation count is selected.
The original implementation utilizes this MCTS formulation and self-play~\citep{silver2017human} to collect state-action pairs (or trajectories).
These state-action pairs are then used to continuously update the policy network with a behavioural cloning algorithm. The three games studied have a reward of $0$ in all but the last time-step, where the reward is $1$ if won, $-1$ if lost and $0$ if the game draws. This 'terminal value' is then paired with all states of the corresponding trajectory and used for value function learning.

MuZero~\citep{schrittwieser2020mastering} is a model-based extension of AlphaZero,
which still relies on MCTS but generalizes to single-player domains and infinite horizon problems with intermediate rewards. While the policy targets stay the same as in AlphaZero,
the value targets need to be adapted for the intermediate rewards. Specifically the return $z_t$ is computed for each state in the trajectory with $z_t = r_{t} + \gamma r_{t+1}  + \gamma^2 r_{t+2}  + \dots + \gamma^n \nu_{t+n} $ where $\nu_t$ is the prediction of the return by the value network for time step $t$. Note, that the prediction $\nu$ can be neglected if the given environment has no infinite horizon but a max reachable time step $t_{max}$ and $n=t_{max}$.

While managing a power grid is not a two-player game like Chess or Go,
it still bears some inherent similarities
when it comes to the long-term effect a single action can have on the development of the grid (``game'')
as well as the exponential nature of the action space (see Section~\ref{subsec:action_space}).
Since AlphaZero has already proven superior performance for problems
where long-term planning in large combinatorial action spaces is required,
we see it as a natural fit for the topology optimization problem in large power grids.
Furthermore, power grid management is a domain
where fast and accurate simulations have been optimized for decades
and are already widely used in practice, e.g., for day-ahead or intra-day operation planning.
The nature of the optimization problem
as well as the fact that power grid load flow simulators are already established in practice
set the stage for our modified version of the AlphaZero algorithm in the single-player infinite horizon power grid setting also utilizing modifications of MuZero.

\begin{figure}
  \centerline{\includegraphics[width=1\textwidth]{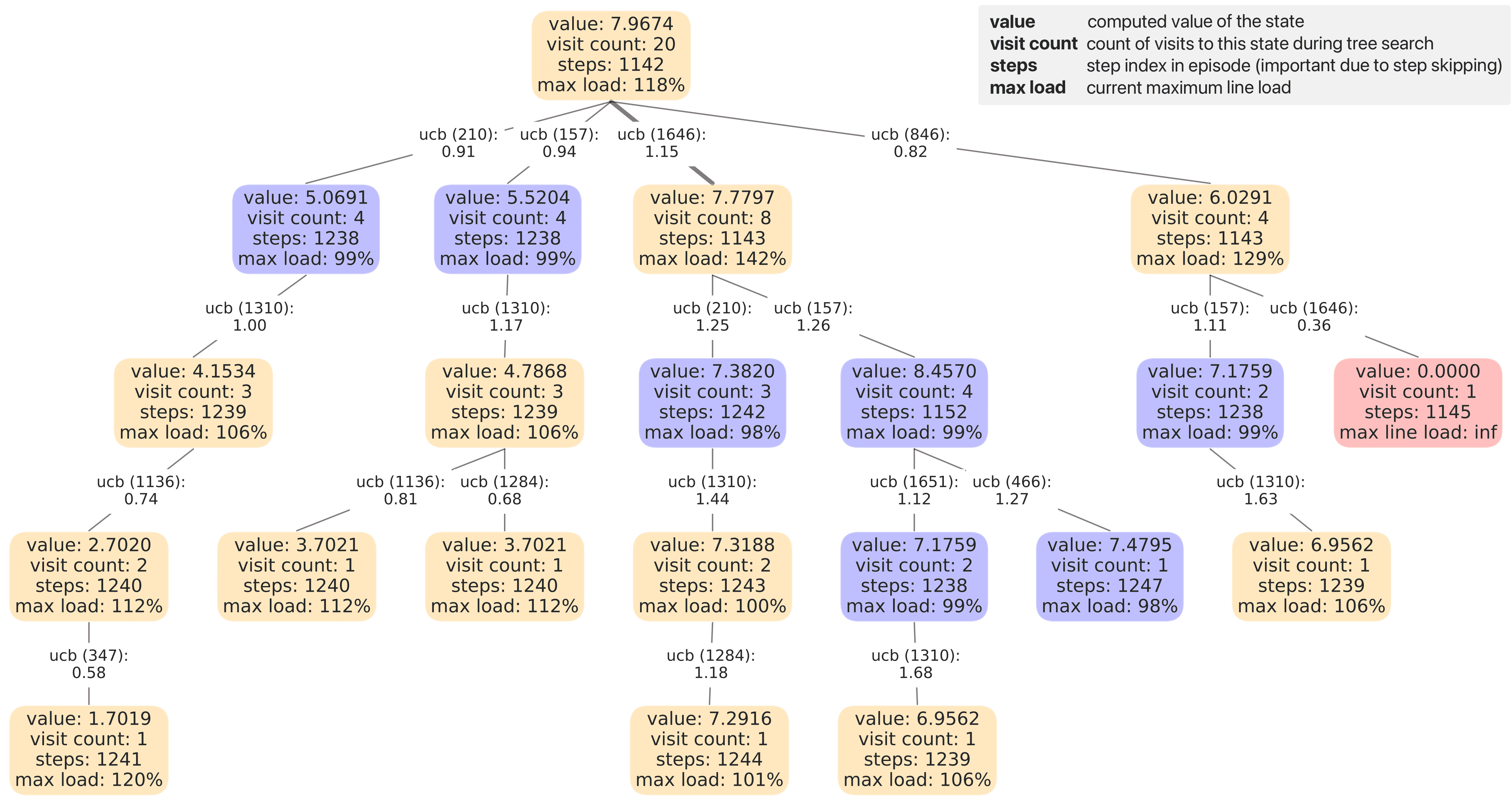}}
  \caption{Example of a grid topology MCTS tree. Nodes represent states  (root is the current state); edges are simulated actions. Yellow nodes are critical states (maximum line load > 98\%), red nodes are black-out states and blue nodes are critical states that internally skipped two or more steps because they were deemed safe by the \emph{Grid State Observer} (see Section~\ref{subsec:env_dynamics}). Edges correspond to topology actions with the action ID (in brackets) and the corresponding upper confidence bound.}
  \label{fig:mcts_tree}
\end{figure}

\subsection{Early Stopping and Action Selection}
%
Considering that AlphaZero is a tabula rasa learning algorithm, an extensive amount of data needs to be collected during the training process, which requires large-scale parallelization and as a consequence a lot of compute resources. For Chess, Shogi and Go~\citet{silver2017alphazero} report using 5,000 first-generation TPUs~\citet{jouppi2017datacenter} to generate self-play games and 64 second-generation TPUs to train the neural networks.
To substantially reduce the amount of resources required to train such a model to only 64 CPU cores to generate trajectories and 1 GPU to train the neural networks, we devised an MCTS early stopping mechanism that terminates the simulation search
once a good (enough) solution has been found.
For our experiments we use two stopping criteria: the number of distinct recovery nodes found in the tree, and the existence of a node that reaches the end of a training episode (i.e., 2016 steps in our case).
Recovery nodes (blue nodes in Figure~\ref{fig:mcts_tree})
represent states that were able to skip at least $t_{skipped}$ environment steps
in a stable grid configuration (i.e., deemed safe by the \emph{Grid State Observer})
before entering a situation where intervention by the agent is required again (i.e., unsafe state).
Once the number of distinct recovery states in the tree exceeds the threshold $t_{stopping}$ the simulation phase is (early-) stopped and the algorithm proceeds to the action selection phase. As an example, if $t_{skipped}=10$ and $t_{stopping}=6$ then the MCTS search with the current tree depicted in Figure~\ref{fig:mcts_tree} would be stopped at this point.
Our experiments in Section~\ref{sec:experiments} show that the effective number of MCTS simulation steps
until the early stopping criterion is hit
steadily decreases throughout training as the prior of the policy network improves. 

Since in many cases MCTS is aborted early due to early stopping,
the action selection based on the visitation count (used in the original algorithm) has to be adjusted.
With the main objective in mind -- keeping the grid up and running
in a stable state for as long as possible (see Section~\ref{subsec:power_grid}) -- we define the action strategy as: choose the action that corresponds to the child node leading to the maximum number of reachable steps in the tree.
In our example in Figure~\ref{fig:mcts_tree}, this corresponds to picking the action with id $1646$.
As~\citet{Browne2012mcts} already suggested in their survey paper
we are also able to improve the efficiency of MCTS by
incorporating domain knowledge into the search process.

\subsection{Heuristic Value Function}
Another modification we make to the AlphaZero algorithm is
the use of an engineered heuristic value function instead of a learned neural network.
This design choice enables us,
similar to early stopping and action selection,
to incorporate domain knowledge into the algorithm
resulting in a speed up (reduced training time) and more stable learning behaviour.
In our experiments learning the value function was inferior
compared to the heuristic value with respect to
the required number of training samples,
final performance of the agent as well as training stability\footnote{
This might only hold in our low resource training regime (1~GPU, 64 CPU cores) compared to the original AlphaZero training setup and might flip when scaling up compute.}.
Additionally, in contrast to Chess, Go or Shogi
it is easier to design a non-sparse reward for the power grid environment.
In general, the value for a node is the accumulated value of the tree branch below
divided by the visitation count of the node.
Whenever a new leaf node is added (expanded) to the MCTS tree
its initial value $\nu$ is predicted using either the value network
or in our case a heuristic value function.
Subsequently, the value gets back-propagated up the tree to the root node using the discounting factor $\gamma$. Averaging the value of a node in such a way further stabilizes the search especially if the reward is intermediate as is the case with the power grid environment.
The heuristic value function for state $S_t$ employed in our experiments
is based on the assumption that the current reward will remain similar for subsequent states.
Although not entirely correct, this assumption is still reasonable
as power grids are not expected to exhibit substantial state changes within a few time steps
except for unforeseeable events, which are also not predictable by a learned value function.
In particular, the heuristic value is defined as the asymptotic value
$\nu_t = \sum_{j=t}^{t+h} \gamma^j r_j $
with some large enough horizon $h$ (depending on $\gamma$)
where $r_j$ is the reward received at time step $j$.

\section{Joint Topology Optimization and Redispatching}
\label{sec:join_topo_redis}

This section describes the redispatching controller
used as a baseline agent in~our experiments,
and how it is combined with our learned topology agent.

\textbf{Redispatching Controller.}
Conceptually, joint optimization is applicable to any redispatcher,
including the mature systems deployed in grid operation environments.
For our experiments, we opted for a runtime-efficient approximation approach,
based on Cross-Entropy (CE) optimization~\citep{boer2005ce}\footnote{
Competitive with the Grid2op's official convex optimization baseline.}.
As a cost function, we simulate the sampled action (see Section~\ref{subsec:observation_space}) and calculate the sum of all overflowing line loads (or take the maximum line load if no overflows remain). 

\textbf{Redispatching vs. Topology Optimization.}
During congestion, we attempt to select the~most suitable remediation measure at~hand.
In cases when a \textit{topology-only action} (from our learned topology agent)
or a \textit{redispatching-only action} (from the redispatching controller) resolves the congestion (in simulation),
it is selected for execution.
If both resolve the congestion (all loads $<98\%$)
the topology action is preferred to save redispatching costs.
In cases when none of these two resolves the congestion, we employ superimposed redispatching. 

\textbf{Superimposed Redispatching.} We expect that not all topology actions are equally well-suited to be combined with redispatching.
Hence, we work with the top-5 topology optimization candidates, re-running the redispatching optimization for each one, taking the proposed topology action into account. This enables the redispatching optimizer to tailor the redispatching action to the proposed topology optimization, and provides a joint topology and redispatch action. We simulate the candidates and pick the action that results in the lowest simulated remaining congestion.

\section{Experimental Evaluation}
\label{sec:experiments}
This section describes agent training,
experimentally evaluates the congestion management potential of the proposed topology optimizer
and tests its interoperability with traditional measures.

\textbf{Experimental Environment.}
Experiments are carried out on the WCCI 2022 power grid
consisting of 118 substations, 186 power lines, 62 generators, 91 loads and 7 storages (see Appendix).
Each episode is a replay of generated loads and generation patterns of 2016 time steps ($\Delta_t=5min$)
resulting in a total time span of seven days.
The environment is designed to also test the ability of agents to establish n-1 grid stability
by randomly disconnecting one out of a predefined set of power lines
at randomly sampled time steps through the episode replays
(also referred to as \emph{opponent attacks}).

\textbf{Unitary Action Reduction.}
The WCCI 2022 grid allows for a total of 72958 topological actions (see~Section~\ref{subsec:action_space}).
Given the size of this action space, it is infeasible to
directly train a policy with AlphaZero and a reasonable amount of compute budget.
Both the output layer of the policy network predicting the prior 
as well as the branching of MCTS nodes would exhibit a dimensionality of 72958.
To mitigate this issues we reduce this original action set
by running a brute force search for the top 2000 most frequent actions~\citep{marot2021retrospective}.
This reduced set is the basis for all topology agents in the experiments below.
Researching more elegant solutions for dealing with the large action space
is a promising open research direction,
which we discuss further in Section~\ref{sec:practical_relevance}.

\begin{figure}[h]
  \centerline{\includegraphics[width=.98\textwidth]{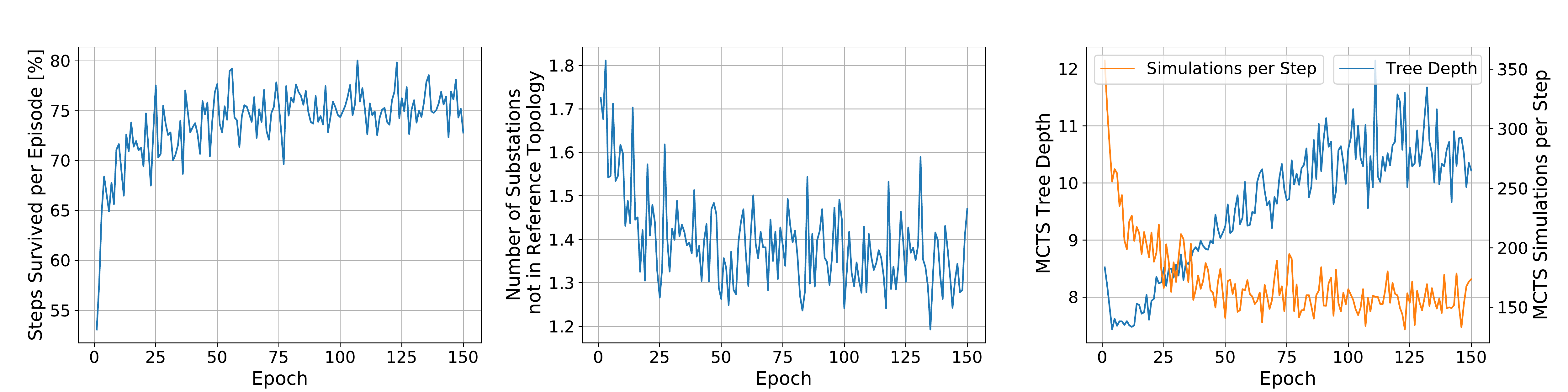}}
  \caption{Evolution of the topology agent during training
  (mean values per epoch).
  }
  \label{fig:training_curves}
\end{figure}

\textbf{Training.} For training our topology policy network with AlphaZero, we have 1662 generated training replays (scenarios) available.
For each training rollout, the random generator of the opponent is seeded differently,
which allows us to increase the effective number of training episodes available.
Figure~\ref{fig:training_curves} shows the evolution of some evaluation measures of the topology agent during training.

We can see that the average ratio of steps survived already improves to over 65\% after the first training epoch,
and continues to increase throughout the entire training process to 75\%.
Interestingly, along with the performance improvement,
we observe a decrease in the number of substations required for relieving congestions,
indicating that the agent discovers more efficient strategies to utilize the available grid topology.
This is also in line with the right-most plot showing a continuous increase of the average MCTS search depth,
allowing the agent to plan and test topological action sequences several time steps into the future.
Note that this increase in the temporal planning horizon takes place even though
the number of simulations required until the early-stopping criteria are met
continues to decrease.
This behavior is a result of the improved accuracy of the learned policy network
because it provides the prior of the PUCT action selection strategy of MCTS.

\textbf{Evaluation.} For evaluation we consider the 52 held out scenarios provided along with the WCCI 2022 grid
and ten different random seeds resulting in 520 distinct evaluation episodes.

As baselines (BL) we consider
\textbf{NoOp (BL)}, an agent taking no grid operations at all,
\textbf{R (BL)}, relying solely on traditional congestion management measures such as redispatching and curtailment 
as well as \textbf{T (brute-force, BL)}, a brute force topology search, 
trying all available actions by testing them with the forecasting simulation
on the current grid state and selecting the one with minimum remaining congestion
(analogous to the action space reduction described above). 

Given the policy network produced by AlphaZero training,
we formulate three different agents  relying solely on topology optimization.
\textbf{T (arg-max)} directly selects the top prediction of the policy network for execution.
\textbf{T (top-25)} simulates the top 25 predictions of the policy network on the current grid state and selects the one with minimum remaining congestion
analogously to the brute force baseline.
\textbf{T (MCTS)} runs a full topology optimization tree search as during AlphaZero training,
utilizing the learned policy network as a prior for the tree policy
along with environment oracle information (i.e., it is able to look into the future).
Hence, it can be seen as a theoretical performance upper bound for topology-based congestion management on the evaluation grid. 

Finally, we define a combined agent \textbf{T (top-5) + R},
testing the interoperability of the trained topology agent with joint redispatching.
This is relevant for two main reasons:
(1) from a grid operation perspective, it is the most relevant setting
with respect to deploying a topology optimization agent in practice, and 
(2) from a RL perspective, it is of interest because redispatching was never applied during training,
and the experiments reveal the robustness of the topology agent with respect to the changed observations (distribution shift).
%
Further details for this agent are provided in Section~\ref{sec:join_topo_redis}. 

\begin{table}[h]
  \caption{Comparison of policies on the 52 WCCI 2022 evaluation episodes and ten different random seeds.
  All measures except \emph{Steps Survived} are averaged by the total steps survived
  to allow for a fair comparison independent of the agent's performance.}
  \label{tab:sample-table}
  \centering
  \begin{tabular}{lcccc}
    \toprule
    Agent      & Steps Survived  & Step Time[ms] & Redispatch[MW] & Curtailment[MW] \\
    \midrule
    NoOp (BL)               & 19.2\% & 8.9 & 0.0 & 0.0 \\
    R (BL)                  & 74.5\% & 31.0 & 504.2 & 484.4 \\
    \midrule
    T (brute-force, BL)     & 61.1\% & 153.3 & 0.0 & 0.0 \\
    T (arg-max)             & 50.4\% & 13.4 & 0.0 & 0.0 \\
    T (top-25)              & 65.3\% & 34.2 & 0.0 & 0.0 \\
    T (MCTS, oracle)        & 76.9\% & 1714.2 & 0.0 & 0.0 \\
    \midrule
    T (top-5) + R           & 82.1\% & 53.3 & 202.8 & 193.4 \\
    \bottomrule
  \end{tabular}
\end{table}
Table~\ref{tab:sample-table} summarizes the performance of the respective agents
and compares them based on the overall steps survived,
the average time required to take one environment step (measured on a single CPU core),
as well as the energy consumed by traditional congestion management measures.
The NoOp baseline maintains the grid for about 19\% of overall steps
and does not introduce any congestion management costs.
R (BL), the agent applying traditional congestion management measures,
maintains the grid for 74.5\% of the steps,
but at the average redispatching and curtailment costs of 988MW per step.
By contrast, the topology-based agents do not introduce any congestion management costs.
While T (brute-force, BL) maintains the grid for 61\% of the steps
the plain policy network T (arg-max) only achieves 50\%.
T (top-25) survives 65\% of the steps and hence outperforms the brute force agent.
This is an interesting results
as it shows that the already reduced action space dimensionality of 2000 still benefits from simulation testing.
More importantly it also shows that it is superior to the greedy brute force search
revealing long-term dependencies between subsequent topological actions.
This observation becomes even more evident
when taking the performance of the oracle MCTS agent into account,
which achieves almost 77\% of steps survived.

T (top-5) + R, the agent combining topology optimization with traditional measures,
achieves 82\% of overall maintained steps,
outperforming all remaining agents
while at the same time reducing the average required redispatching power to 396.2 MW;
only 40\% of the original baseline costs.
This is a promising finding as it confirms that
topology optimization is indeed complementary to traditional congestion management measures
but more importantly bears a great potential for
stabilizing the grid and reducing cost while at the same time reducing emissions.
%
%
A closer look at the selected actions additionally
shows that the top topology action candidate
is not always the one most compatible with superimposed redispatching (only in 65\% of the cases).
This suggests that incorporating redispatching already in the training process of the topology agent
is another promising research direction.

\section{Conclusions and Practical Relevance for the Real Power Grid}
\label{sec:practical_relevance}
We propose an AlphaZero-based power grid topology agent
and show how to combine it with traditional congestion management measures.
This joint agent improves the grid resilience
while at the same time reducing costly measures to only 40\% of the original costs.
In the last section of this paper, we
emphasize the practical relevance of the proposed agents with the example of two real world use cases,
and discuss open research problems.

\textbf{Use Case 1: Day-Ahead Planning.}
A common practice in real-world grid operation is day-ahead planning,
where a congestion management plan is created for the~entire day ahead,
based on the market outcomes, available demand and renewable generation forecasts,
and a load flow simulation of the real grid.
Both the availability of forecasts and of the load flow simulator
make the~day-ahead planning routine well suited for RL-based power grid operation planning.

\textbf{Use Case 2: AI-Assisted Control.}
 AI-assisted control addresses the exploration problem by utilizing an RL-based agent for generating remediation action candidates. These are then evaluated by a human (or any conventional process) in order to maintain the strict safety and interpretability requirements that power grids, as critical infrastructure, place on all implemented measures.
 We present a web application that demonstrates an elementary version of AI-assisted control with multi-candidate, multi-step congestion remediation recommendations on a simulated power grid\footnote{Please see the Appendix for more information, or visit our demo at:
 \url{http://grid-demo.enlite.ai/}
 }.

\textbf{Open Problem 1: Redispatching Integration.}
Our experimental results reveal that the efficient and effective combination of redispatching with topology optimization is a promising open research direction.
In practice, costly congestion management measures are usually favored over topology corrections.
At the same time, our approach shows that joint actions have a strong potential. However, the current setup does not yet allow for more coherent and co-operative joint actions as it is, e.g., infeasible to train our topology optimization agent together with existing redispatching controllers, due to computational limitations.
Exploring ways for computationally more efficient redispatching would open the door
to combining the two paradigms already during training.
This would allow the topology agent to adjust its control strategies with respect to the redispatcher\footnote{
Our initial attempts to approximate redispatching with behavioural cloning have not been successful.}.

\textbf{Open Problem 2: Action Representation.}
To help mitigate the exploration problem, agents have so far been either developed and tested on moderately sized power grids,
still exhibiting a~tractable exploration space, or,
as in our case, utilize only a drastically reduced set of available topology actions.
We think that devising improved, potentially domain-knowledge informed action space formulations would allow the agent to better leverage the full optimization potential of the underlying power grid.

\textbf{Conclusions.} Given the promising results, we see RL-based optimization
as a promising avenue for unlocking the potential of topology-based congestion management
and the efficient utilization of renewable energy generation, contributing to the societal goal of decarbonizing the energy sector.

\medskip

\begin{ack}
We would like to thank RTE France for developing and providing the Grid2Op environment
enabling the line of RL(Agent)-based research on power grids.
We would also like to thank Antoine Marot (RTE France) and Jan Viebahn (TenneT) for useful discussions
and Benjamin Donnot (RTE France) for his fast \textit{lightsim2grid} loadflow simulation backend.
\end{ack}

\small
\bibliographystyle{authordate1}
\bibliography{main.bib}

\clearpage
\appendix

\section*{APPENDIX}

\subsection*{Default Topology of the WCCI 2022 Grid}
\begin{figure}[h]
  \centerline{\includegraphics[width=1.37\textwidth, angle=90]{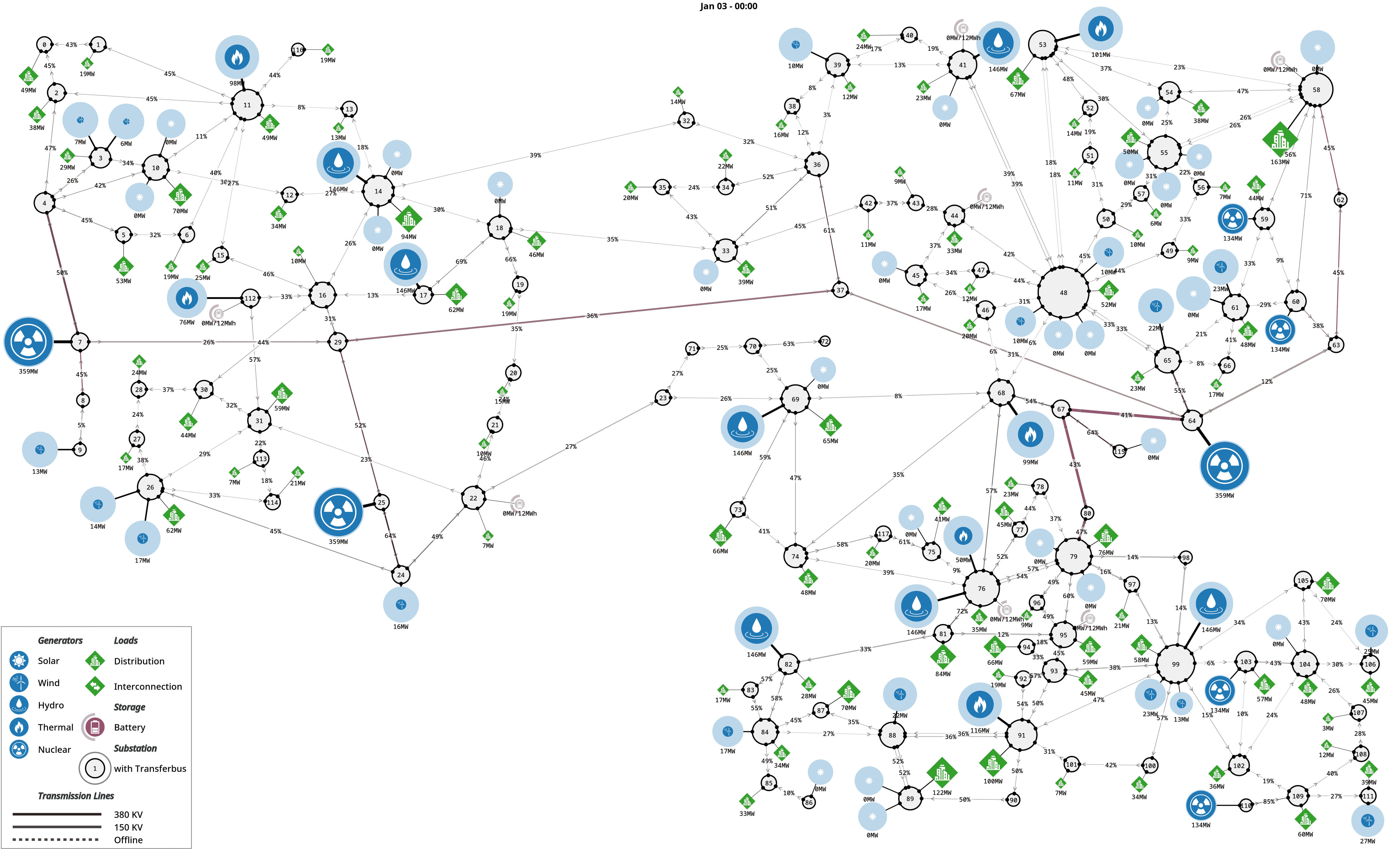}}
  \caption{The WCCI 2022 power grid including all relevant components.}
  \label{fig:wcci_2022_grid_topology}
\end{figure}

Figure~\ref{fig:wcci_2022_grid_topology} shows the WCCI 2022 grid used in our experimental evaluation. It consists of 118 substations, 186 power lines, 62 generators, 91 loads and 7 storages.
The theoretical upper bound of possible topology configurations for this grid is $2^{186 \cdot 2 + 62 + 91 + 7}=2^{532}$.

\subsection*{AI-Assisted Control: Real-Time Remedial Action Recommendation}
\begin{figure}[h]
  \centerline{\includegraphics[width=.99\textwidth]{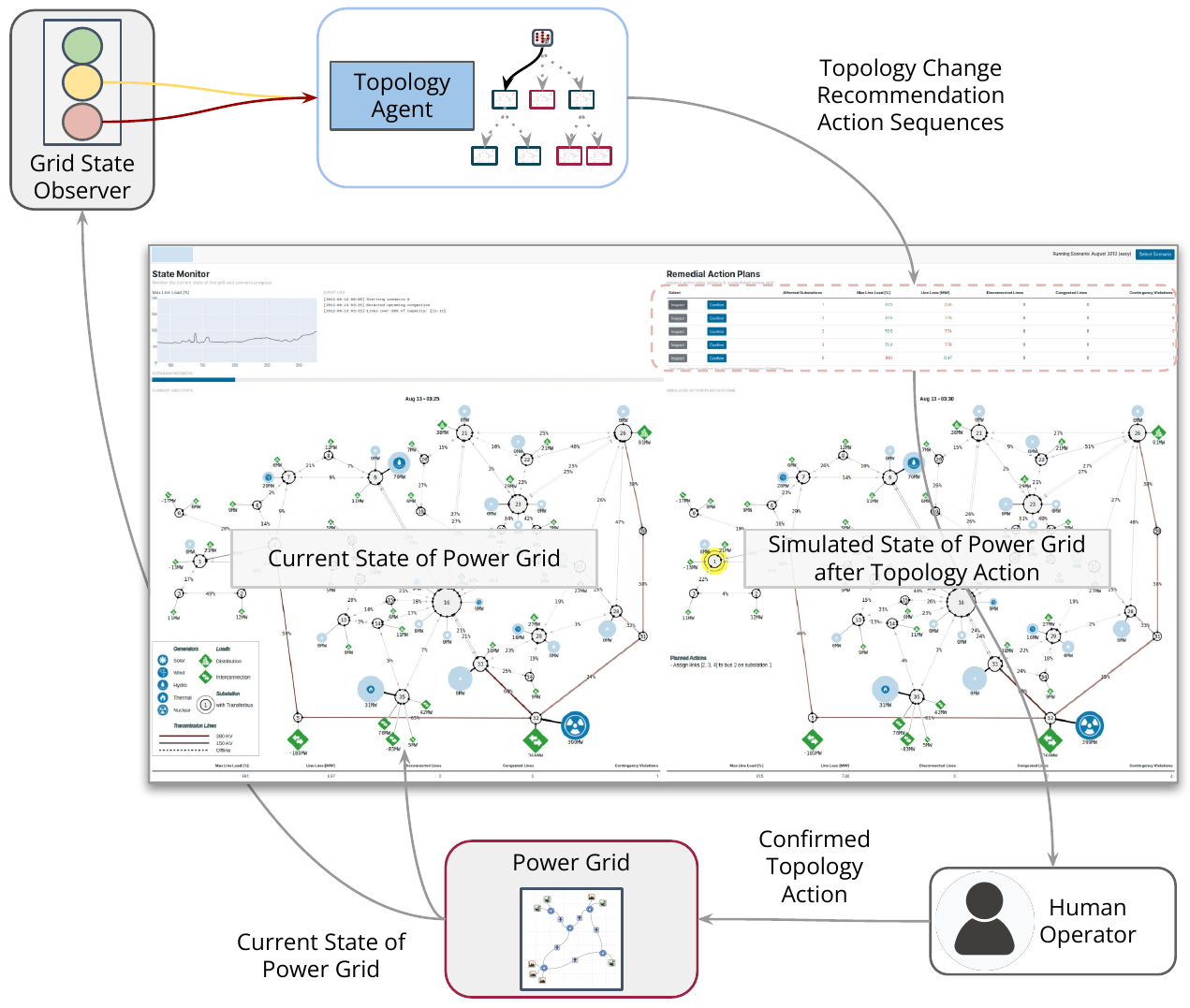}}
  \caption{Overview of AlphaZero-powered real-time remedial action recommendation assistant.}
  \label{fig:assistant_overview}
\end{figure}

Figure~\ref{fig:assistant_overview} shows an overview of our real-time remedial action recommendation assistant demo
(
\url{http://grid-demo.enlite.ai/}
)
as a concrete example for \emph{human-in-the-loop} decision making.

The central design principle of our AI assistant is
to support and enhance human decision making
by recommending viable action scenarios along with augmenting information
explaining how these recommendations will most likely turn out in the productive system.
The final decision is left to the human operator to preserve human control.
The working principle of the assistant is as follows:
\begin{itemize}
    \item The \emph{Grid State Observer} continuously monitors the current state of the productive power grid.
    \item Once a non-safe state is encountered the agent starts a policy network guided tree search to discover a set of topology change action scenarios that are capable of recovering from this critical situation (e.g., relieving the congestion).
    \item A ranked list of topology change candidates -- the top results of the tree search -- is presented to the human operator in a graphical user interface for evaluation (testing the impact of a action candidate in a load flow simulation) and selection. Along with the potential for relieving the congestion other grid specific safety considerations are taken into account (e.g., a contingency analysis for n-1 stability of the respective resulting states). 
    \item The human operator evaluates the provided set of suggested topology change actions.
    \item Once satisfied, he/she confirms the best action candidate for execution on the productive power grid.
    \item The selected action candidate is applied to the power grid and the resulting state is again fed into the \emph{Grid State Observer} and visualized for human operators in the graphical user interface. This closes the human in the loop workflow.
\end{itemize}

When operating a real power grid, most congestion is anticipated and addressed in advance to ensure safe operating conditions at all times. The time window in which a remediation measure has to be implemented, to prevent the grid from becoming unstable, varies from several hours to a few minutes in the event of unexpected fluctuations. Our AI assistant demonstrates that recommendations can be produced in less than one second, meaning human operators can incorporate agent’s outputs into their evaluation procedures without loosing valuable time (see Table \ref{tab:sample-table}, where our proposed agent is denoted as \textit{T (top-5) + R}).

\subsection*{Agent Environment Interaction Overview}
\begin{figure}[h]
  \centerline{\includegraphics[width=.89\textwidth]{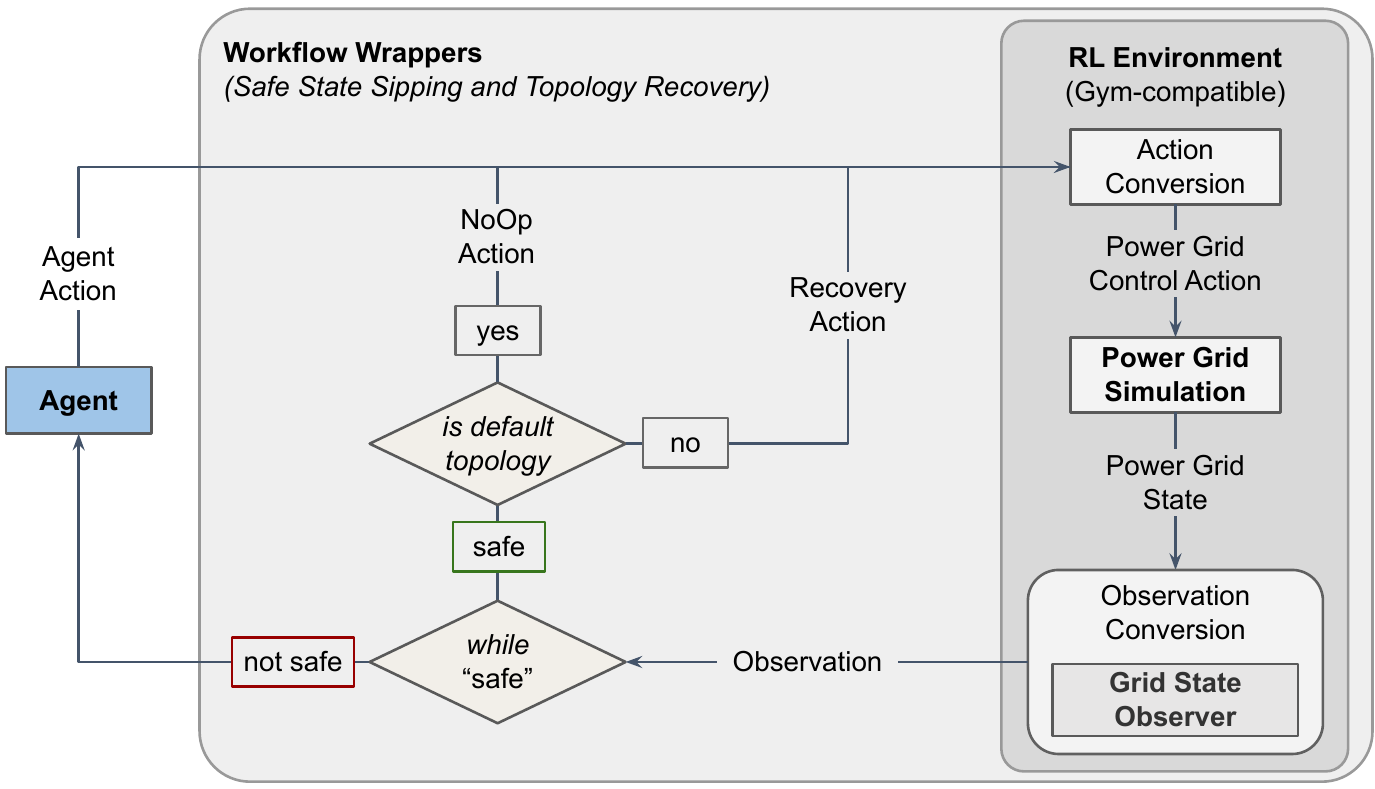}}
  \caption{Overview of the agent environment interaction workflow.}
  \label{fig:workflow_wrappers}
\end{figure}

Figure~\ref{fig:workflow_wrappers} provides an overflow of the employed agent environment interaction workflow
described in Section~\ref{subsec:env_dynamics}.
Involved components are the agent, the power grid load flow simulator wrapped into an RL environment
as well as safe state skipping and topology recovery.

\subsection*{Reward Formulation}

This section describes the reward formulation for training the topology agent.
We utilize a shaped reward based on the cumulative sum of all overflowing line loads, which the agent aims to minimize.

To formulate the reward, we first calculate the coefficient $u$, which summarizes the (overflowing) line loads. 

If $\rho_{\textrm{max}} \le 1$, i.e., there is currently no overflow, and line loads of all lines are within the allowed bounds, $u$ is calculated as

\begin{equation}
    u = \textrm{max}(\rho_{\textrm{max}} - 0.5, 0)
\end{equation}

where $\rho_{\textrm{max}}$ is the current maximum line load.

In case of an overflow, i.e., when $\rho_{\textrm{max}} > 1$, $u$ is computed as

\begin{equation}
    u = \sum_i ( \rho_i - 0.5 ) \quad \textrm{for} \quad i \in \{ [1, n]: \rho_i > 1 \}
\end{equation}

where $n$ is the number of power lines in the grid, $i$ are indices of currently overflowing power lines, and $\rho_i$ denotes the current load of the (overflowing) power line $i$.

\begin{figure}[h]
  \centerline{\includegraphics[width=.95\textwidth]{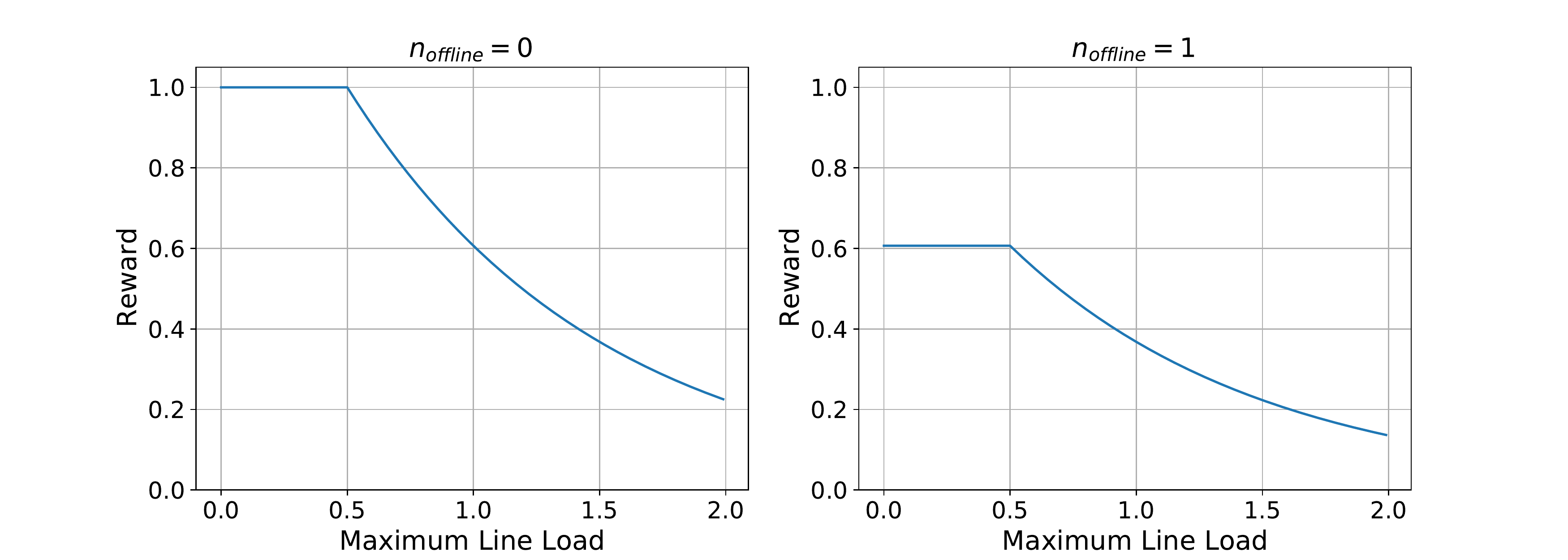}}
  \caption{Behaviour of shaped reward with respect to line loads and offline power lines.}
  \label{fig:shaped_reward}
\end{figure}

Then, utilizing $u$ calculated above, we take into account offline lines and apply exponential decay to~obtain the~shaped~reward $r$ as

\begin{equation}
    r = \exp \left( - u - 0.5 \cdot n_{\textrm{offline}} \right) 
\end{equation}

where $n_{\textit{offline}}$ is the number of lines which are currently offline as a result of an overflow or agent's actions (i.e., we do not consider lines that are offline because of maintenance or opponent attacks). Figure~\ref{fig:shaped_reward} shows two examples how the reward behaves with respect to current line loads and the number of offline lines.

\subsection*{Action Selection Overview}
\begin{figure}[h]
  \centerline{\includegraphics[width=.98\textwidth]{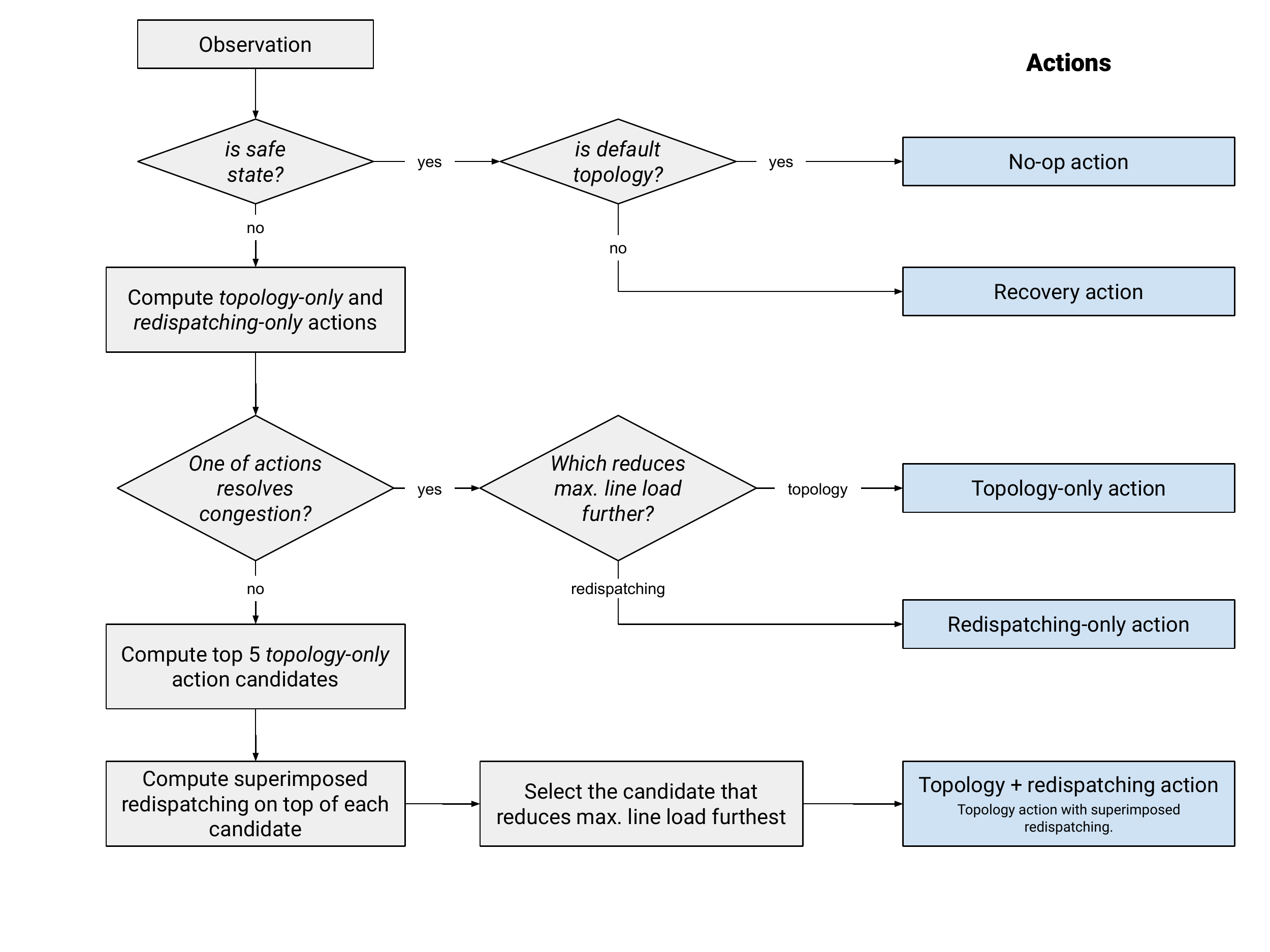}}
  \caption{Overview of the action selection mechanism.}
  \label{fig:action_selection_flow}
\end{figure}

Figure~\ref{fig:action_selection_flow} provides a more detailed overview of the action selection flow (already partially captured in Figure~\ref{fig:workflow_wrappers}), as described in Sections~\ref{subsec:env_dynamics} and~\ref{sec:join_topo_redis}.
This action selection mechanism is implemented through the safe-state skipping module, topology recovery module, topology agent and redispatching controller.

\newpage

\subsection*{Topology Action Sequence}

\begin{figure}[h]
  \centerline{\includegraphics[width=1\textwidth]{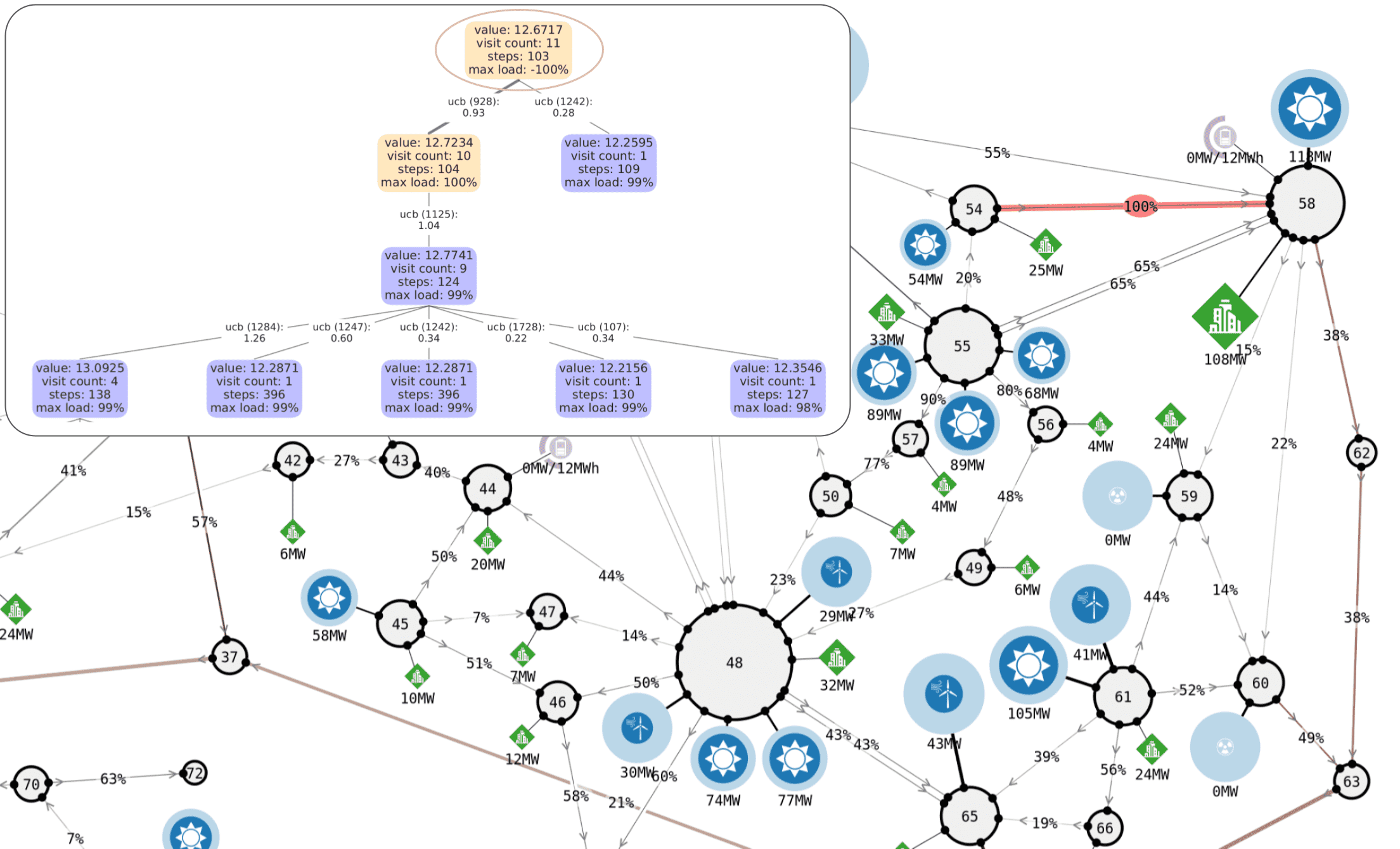}}
  \caption{$t=103$, Initial congestion on the power line between substation 54 and 58.}
  \label{fig:mcts_tree_0}
\end{figure}

\begin{figure}[h]
  \centerline{\includegraphics[width=1\textwidth]{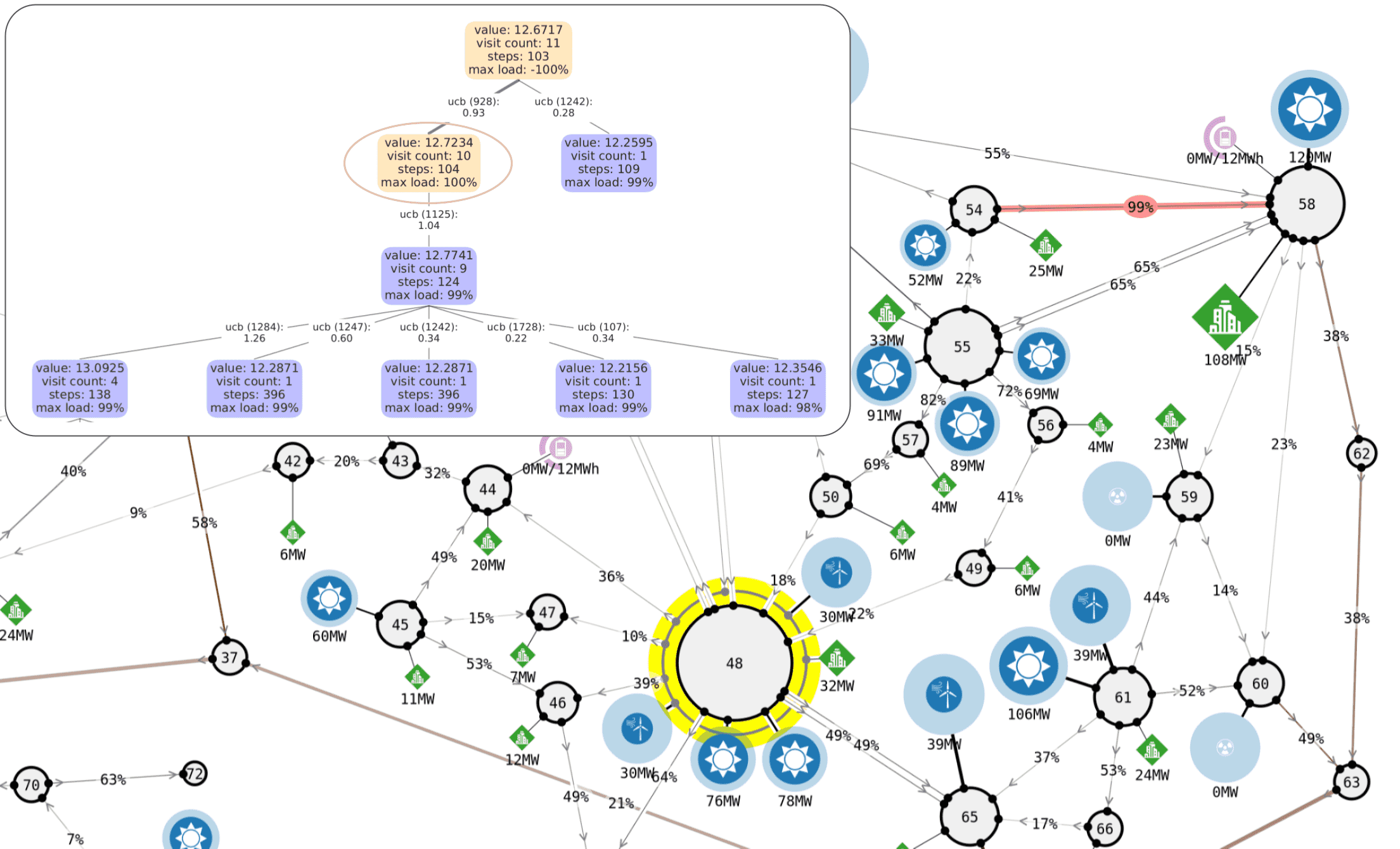}}
  \caption{$t=104$, First step of topology action sequence performing a bus split on substation 48.}
  \label{fig:mcts_tree_1}
\end{figure}

\begin{figure}[t!]
  \centerline{\includegraphics[width=1\textwidth]{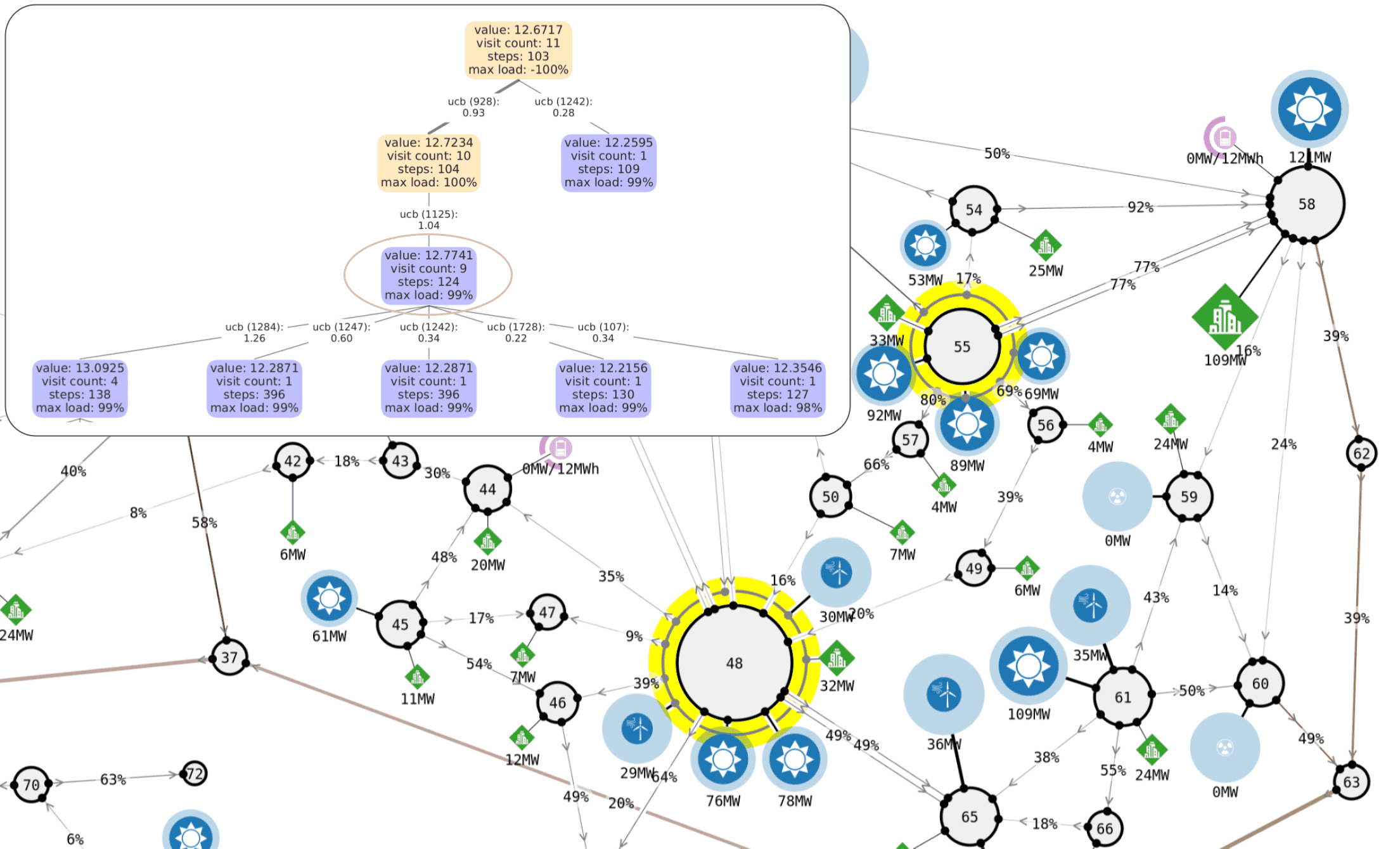}}
  \caption{$t=105$, Second step of topology action sequence performing a bus split on substation 55.}
  \label{fig:mcts_tree_2}
\end{figure}

Figures~\ref{fig:mcts_tree_0} to~\ref{fig:mcts_tree_2} show a topological action sequence example where a grid congestion
on the power line between substation 54 and 58 is resolved by bus splits on substation 48 and 55 without applying any traditional measures.

\end{document}